\documentclass[utf8]{owm_class} 

\usepackage{url,lineno,microtype,subcaption}
\usepackage[onehalfspacing]{setspace}
\usepackage{xr}
\usepackage{changes}
\usepackage{xcolor,soul}
\definechangesauthor[name={Chen Yang}, color=red]{CY}
\def\firstAuthorLast{ } 
\def\Authors{Guanxiong Zeng\,$^{1,2,*}$, Yang Chen\,$^{1,*}$, Bo Cui\,$^{1,2}$ and  Shan Yu\,$^{1,2,3,\dagger}$}

\def\Address{$^{1}$Brainnetome Center and National Laboratory of Pattern Recognition, Institute of Automation, Chinese Academy of Sciences, 100190 Beijing, China.\\
$^{2}$University of Chinese Academy of Sciences, 100049 Beijing, China.\\
$^{3}$Center for Excellence in Brain Science and Intelligence Technology,
Chinese Academy of Sciences, 100190 Beijing, China.
}

\def\corrAuthor{}
\def\corrEmail{\textcolor[rgb]{0.00,0.00,1.00}{shan.yu@nlpr.ia.ac.cn}}

\begin{document}

\onecolumn
\firstpage{1}

\title[]{{Continual Learning of Context-dependent Processing in Neural Networks}}

\author[\firstAuthorLast ]{\Authors} 
\address{} 
\correspondance{} 

\extraAuth{}

\maketitle


\begin{abstract}
	
Deep neural networks (DNNs) are powerful tools in learning sophisticated but fixed mapping rules between inputs and outputs, thereby limiting their application in more complex and dynamic situations in which the mapping rules are not kept the same but changing according to different contexts. To lift such limits, we developed a novel approach involving a learning algorithm, called orthogonal weights modification (OWM), with the addition of a context-dependent processing (CDP) module. We demonstrated that with OWM to overcome the problem of catastrophic forgetting, and the CDP module to learn how to reuse a feature representation and a classifier for different contexts, a single network can acquire numerous context-dependent mapping rules in an online and continual manner, with as few as $\sim$10 samples to learn each. This should enable highly compact systems to gradually learn myriad regularities of the real world and eventually behave appropriately within it.	
	
\end{abstract}

\section*{Introduction}

One of the hallmarks of high-level intelligence is flexibility \cite{Newell1994}. Humans and non-human priamtes can respond differently to the same stimulus under different contexts, e.g., different goals, environments, and internal states \cite{Miller1951,Desimone1995,Mante2013,Siegel2015}. Such an ability, named cognitive control, enables us to dynamically map sensory inputs to different actions in a context-dependent way \cite{Miller1999,Wise1996,Passingham1993}, thereby allowing primates to behave appropriately in an unlimited number of situations with limited behavioral repertoire\cite{Miller2001,Miller2000}. However, this flexible, context-dependent processing is quite different to that found in current artificial deep neural networks (DNNs). DNNs are very powerful in extracting high-level features from raw sensory data and learning sophisticated mapping rules for pattern detection, recognition, and classification \cite{LeCun2015}. In most networks, however, the outputs are largely dictated by sensory inputs, exhibiting stereotyped input-output mappings that are usually fixed once training is complete. Therefore, current DNNs lack sufficient flexibility to work in complex situations in which 1) the mapping rules change according to context and 2) these rules need to be learned sequentially when encountered  from a small number of learning trials. This constitutes a significant gap in the abilities between current DNNs and primate brains.
	
In the present study, we propose an approach, including an orthogonal weight modification (OWM) algorithm and a context-dependent processing (CDP) module, that enables a neural network to progressively learn various mapping rules in a context-dependent way. We demonstrate that with OWM to protect previously acquired knowledge, the networks could sequentially learn up to thousands of different mapping rules without interference, and needing as few as $\sim$10 samples to learn each.  In addition, by using the CDP module to enable contextual information to modulate the representation of sensory features, a network can learn different, context-specific mappings for even identical stimuli. Taken together, our proposed approach can teach a single network numerous context-dependent mapping rules in an online and continual manner.

\section{Orthogonal Weights Modification (OWM)} 
\begin{figure}[!h]
  \centering
  \includegraphics[width=1.0\textwidth] {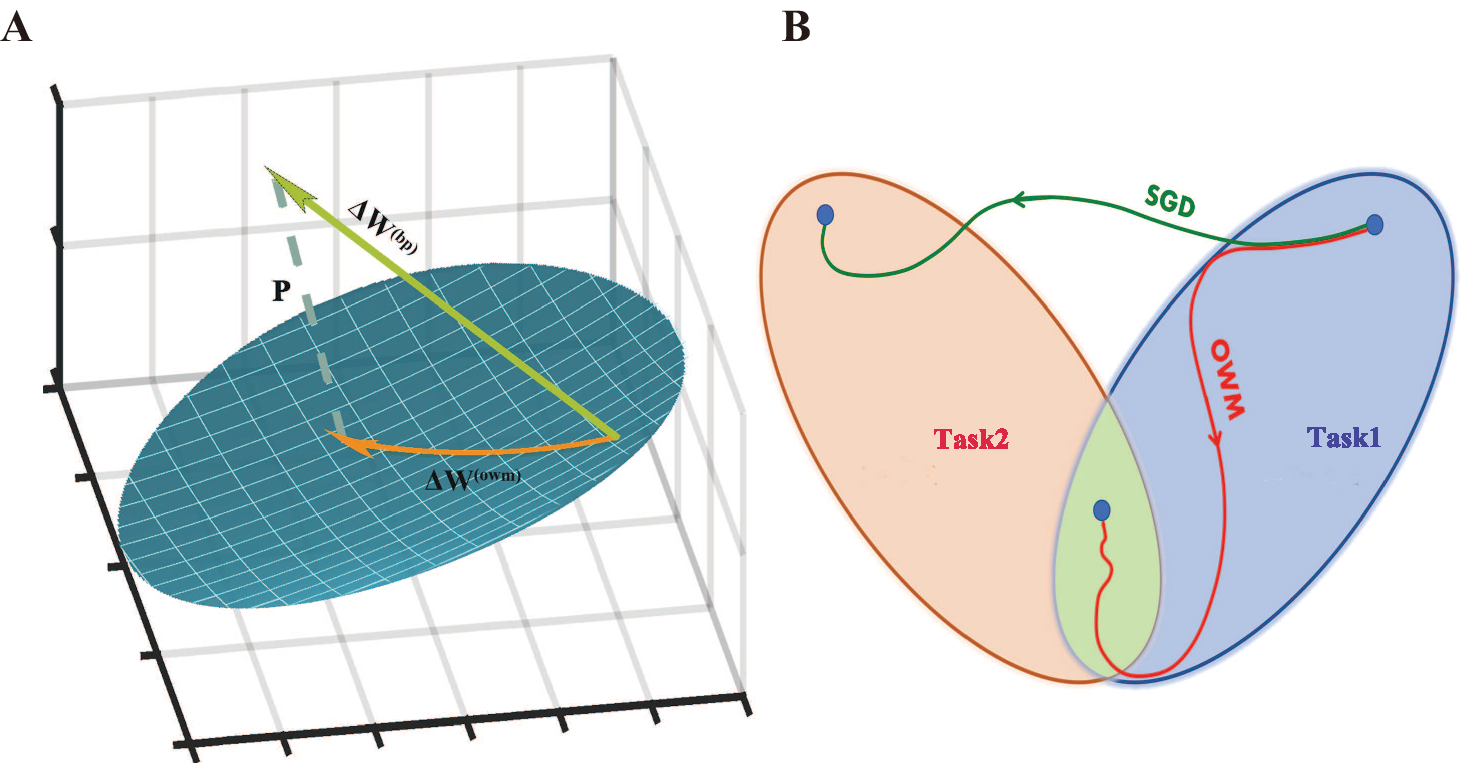}
  \caption{\textbf{Schematic diagram of OWM}. \textbf{a}, In the new task training process, the original weight modification calculated by the standard backpropagation (BP),  $\Delta {\bf{W}^{BP}}$, is projected to the subspace (dark green surface), in which good performance for learned tasks has been achieved. As a result, the actual implemented weight modification is $\Delta {\bf{W}^{OWM}}$. This process ensures that the weights configuration after learning the new task is still within the same subspace. 
  \textbf{b}, With the OWM, the training process searches for configurations that can accomplish Task 2 ( pale red area), within the subspace that enables the network to accomplish Task 1 ( blue area). A successful search necessarily stops at a position inside the overlapping subspace ( light green area). In comparison, the solution obtained by stochastic gradient descent search (SGD) is more likely to end outside this overlapping area. }
  \label{Schematic diagram of OWM}
\end{figure}

The first step towards flexible context-dependent processing is to incorporate efficient and scalable continual learning, i.e., learning different mappings sequentially, one at a time. Such an ability is crucial to humans as well as artificial intelligence agents for two reasons: 1) there are too many possible contexts to learn concurrently, and 2) useful mappings cannot be pre-determined but must be learned when corresponding contexts are encountered. 
The main obstacle to achieve continual learning is that conventional neural network models suffer from catastrophic forgetting, i.e., training a model with new tasks interferes with previously learned knowledge and leads to significantly decreases on the performance of previously learned tasks \cite{McCloskey1989,Ratcliff1990,Goodfellow2013,Parisi2018}. 
To avoid catastrophic forgetting, we developed the OWM method. Specifically, when training a network for new tasks, its weights can only be modified in the direction orthogonal to the subspace spanned by all previously learned inputs (termed the input space hereafter) (Fig.~\ref{Schematic diagram of OWM}a and Supplementary Fig. 1). This ensures that new learning processes do not interfere with previously learned tasks, as weight changes in the network as a whole do not interact with old inputs. Consequently, combined with a gradient descent-based search, the OWM helps the network to find a weight configuration that can accomplish new tasks while ensuring the performance of learned tasks remains unchanged (Fig.~\ref{Schematic diagram of OWM}b). This is achieved by first constructing a projector used to find the direction orthogonal to the input space: $\bf{P} = \bf{I} - \bf{A}{\left( {{\bf{A}^T}\bf{A} + \alpha \bf{I}} \right)^{ - 1}}\bf{A}$, where matrix $\bf{A}$ consists of all previously trained input vectors as its columns $\bf{A} = [{\bf{x}_1}, \cdots ,{\bf{x}_n}]$ and $\bf{I}$ is a unit matrix multiplied by a relatively small constant $\alpha$. The learning-induced modification of weights is then determined by $\Delta \bf{W} = \kappa P\Delta {\bf{W}^{BP}}$, where $\kappa$ is the learning rate and $\Delta {\bf{W}^{BP}}$ is the weights adjustment calculated according to the standard backpropagation.  To calculate $\bf{P}$, an iterative method can be used (see Methods). Thus, the algorithm does not need to store all previous inputs $\bf{A}$. Instead, only the current inputs and projector for the last task are needed. This iterative method is related to the Recursive Least Square (RLS) algorithm \cite{Haykin2008,Golub2012} (see Supplementary Information for the discussion), which can be used to train feedforward and recurrent neural networks to achieve fast convergence \cite{Singhal1989,Shah1992}, tame chaotic activities \cite{Sussillo2009} and avoid interference between consecutively loaded patterns or tasks \cite{Jaeger2014,He2018overcoming}.

\begin{figure}[!htbp]
	\centering
	\includegraphics[width=1.0\textwidth] {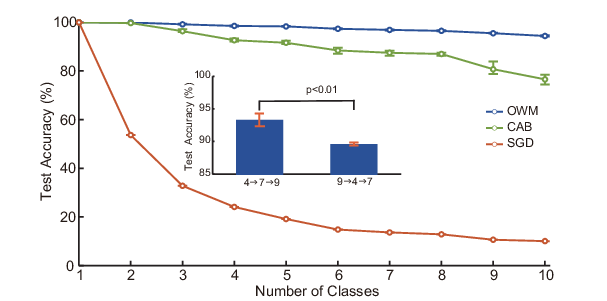} 
	\caption{\textbf{Performance of OWM, CAB, and SGD in 10-disjoint MNIST task.}
	 Test accuracy was plotted as a function of number of classes learned. Results are presented as mean $\pm$ s.d.. For OWM-trained task, the sequence of learnt digits influenced recognition accuracy for specific classes. Inset:  performance of recognizing digit “9” was significantly higher after learning digits “7” and “4”; two-sided \textit{t}-test was applied to assess statistical significance.}
	\label{mnist_Disjoint_Task10}
\end{figure}

We first tested the performance of the OWM on several benchmark tasks of continual learning. Shuffled and disjoint MNIST experiments, in which different tasks involving recognition of handwritten digits need to be learned sequentially  (see Methods and Supplementary Information for details regarding the datasets used in this study), were conducted on the feedforward  network with the rectified linear unit (ReLU) \cite{Nair2010Rectified}.  The OWM was used to train the entire multi-layer networks. For 3- or 10-task shuffled and 2-task disjoint experiments,  OWM resulted in either superior or equal performance in comparison to other continual learning methods without storage of previous task samples or dynamically adding new nodes to the network  \cite{Kirkpatricka2017,Lee2017,Zenke2017,He2018overcoming} (Tables ~\ref{performance of Shuffled MNIST}, ~\ref{performance of Disjoint MNIST}). \textcolor{black}{In the more challenging 10-task disjoint and 100-task shuffled experiments, OWM exhibited significant performance improvement over other methods (Fig. \ref{mnist_Disjoint_Task10} and Table \ref{performance of Shuffled MNIST}).} \textcolor{black}{Interestingly,  for the more difficult continual learning tasks, we found that the order of tasks mattered. As the performance for specific classes can be significantly influenced by the classes learned previously (Fig. \ref{mnist_Disjoint_Task10} inset), suggesting that curriculum learning is a potentially important factor to consider in continual learning.}

\begin{figure}[!h]
  \centering
  \includegraphics[width=1.0\textwidth] {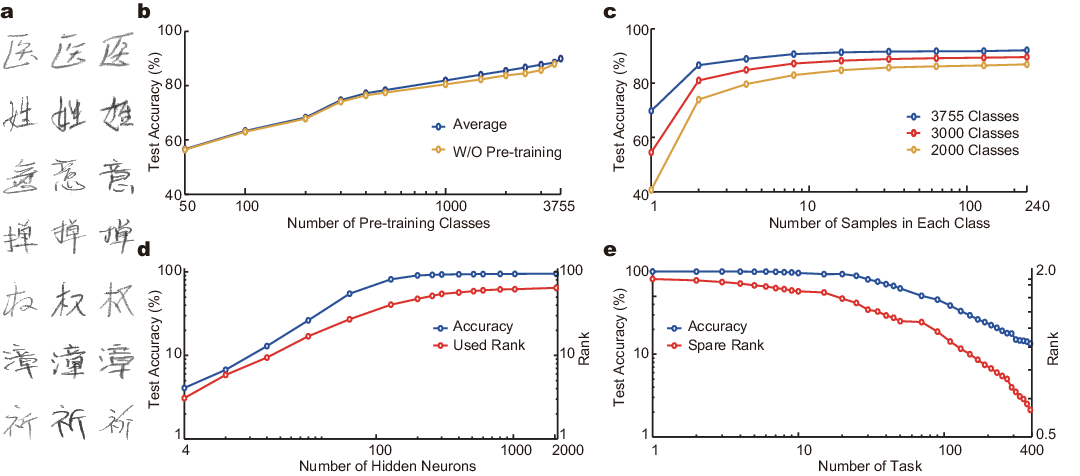} 
  \caption{ \textbf{Continual learning with small sample size achieved by OWM in recognizing Chinese characters.} 
  \textbf{a}, Examples showing seven characters with five samples for each.  
  \textbf{b}, Classification accuracy is plotted as a function of the number of classes used for pre-training the feature extractor. Performance was assessed based on classifying all characters (blue) or characters not included in pre-training (orange).  Variance of test accuracy across classes in each case is reported in Supplementary Table 2.  
  \textbf{c}, Classification accuracy is plotted as a function of sample size used for sequential training, obtained with feature extractors having different degrees of pre-training (color-coded). \textcolor{black}{Performances differed significantly (paired t-test, $p<0.001$) across different degrees of pre-training} (see Supplementary Table  3 for variance in performance across all classes).
  (\textbf{d-e}), Relationship between network capacity for continual learning and rank of the orthogonal projector. In d, the task was to learn 100 classes of Chinese characters sequentially. Average accuracy achieved by the network (blue) and the corresponding value of ($rank(\beta\mathbf{I})-rank(\mathbf{P})$) (red) are plotted with respect to the number of neurons in the hidden layer. 
  In e, the same neural network with 50 neurons in the hidden layer was trained to recognize an increasing number of Chinese characters. 
  Average accuracy (blue) achieved by the network and the corresponding value (red) of $rank_{tot}$ (see Methods) are plotted with respect to the number of tasks/characters.
 }
  \label{OWM in CHW}
\end{figure}

To examine whether the OWM is scalable, i.e., whether it can be applied to learn more sophisticated tasks, regarding both number of different mappings and complexity of inputs, we tested the network's ability in learning to classify thousands of hand-written Chinese characters (CASIA-HWDB1.1) and natural images (ImageNet). The Chinese character recognition task included a total of 3,755 characters forming the level I vocabulary, which constitutes more than $99\%$ of the usage frequency in written Chinese literature \cite{Liu2010} (see Fig.~\ref{OWM in CHW}a for exemplars of characters). In this task, a feature extractor was pre-trained to analyze the raw images. The feature vectors were fed into an OWM-trained classifier to learn the mapping between combinations of features and the labels of individual classes. We found that a classifier trained with the OWM could learn to recognize all 3,755 characters sequentially, with a final accuracy $\sim92\%$ closely approaching the results obtained in human performance when recognizing handwritten Chinese characters ($\sim96\%$) \cite{Yin2013}. Considering humans learn these characters over years and the learning necessarily contains revision, these results suggest that our method endows neural networks with a strong capability to continually learn new mappings between sensory features and class labels.  Similar results were obtained with the ImageNet dataset, where the classifier trained by the OWM combined with a pre-trained feature extractor, was able to learn 1000 classes of natural images sequentially (Supplementary Table 1), with the final accuracy approaching the results obtained by training the system to classify all categories concurrently. These results suggest that, by using the OWM, the performance of the system in classification approached the limit set by the front-end feature extractor, with liability to the classifier caused by sequential learning itself effectively mitigated.    

 \begin{table}[!h]
  \centering
   \begin{tabular}{c|c|c|c|c|c}
   \multicolumn{6}{c}{\textbf{Shuffled MNIST Experiment}} \\
   \hline
   \textbf{\textbf{3 tasks}} & \textbf{Accuracy (\%)} &\textbf{\textbf{10 tasks}} & \textbf{Accuracy (\%)}
   &\textbf{\textbf{100 tasks}} & \textbf{Accuracy (\%)}\\
   \hline
   SGD$^{\#}$\cite{Goodfellow2013} & $71.32\pm1.54^{*}$ & EWC$^{\#}$\cite{Kirkpatricka2017} & $\sim97.0$ 
  &EWC$^{\ddagger}$\cite{masse2018alleviating} & $\sim70.8$\\
   IMM$^{\#}$\cite{Lee2017} & $98.30\pm0.08^{n.s}$ & \textbf{OWM}$^{\#}$ & $97.52\pm0.03$ &
   SI$^{\ddagger}$\cite{masse2018alleviating} & $\sim82.3$ \\
   EWC$^{\#}$\cite{Kirkpatricka2017} & $\sim98.2$ & EWC$^{\dagger}$\cite{He2018overcoming} & $\sim89.0$
   & \textbf{OWM}$^{\ddagger}$ & $\sim85.4$ \\
   \textbf{OWM}$^{\#}$ &  $98.34\pm0.02$ & CAB$^{\dagger}$\cite{He2018overcoming} & $\sim95.2$ 	& \\
   &                           & OWM$^{\dagger}$ & $95.15\pm0.08$ 	& \\
   &                           & SI$^{\ddagger}$\cite{Zenke2017,masse2018alleviating} & $\sim97.0$	& \\
   &                           & \textbf{OWM}$^{\ddagger}$ & $97.64\pm0.03$ 	&\\
   \hline
   \end{tabular}
   \caption{\textbf{Comparison of performance of different methods in Shuffled MNIST task.}
   Network size: $^{\dagger}$, 3-layer networks with [784-100-10] neurons; $^{\#}$, 4-layer networks with [784-800-800-10] neurons; $^{\ddagger}$, 4-layer networks with [784-2000-2000-10] neurons. Results from other methods were adopted from corresponding publications. Results for OWM are represented as mean $\pm$ s.d..  $^{*}$, $p<0.01$. n.s, not significant. EWC: Elastic Weight Consolidation; IMM:Incremental Moment Matching; SI: Synaptic Intelligence}
   \label{performance of Shuffled MNIST}
\end{table}

\begin{table}[!h]
	\centering
	\begin{tabular}{c|c}
	\multicolumn{2}{c}{\textbf{Disjoint MNIST Experiment}} \\
	\hline
	\textbf{Methods} & \textbf{Accuracy (\%)}\\
	\hline
	EWC$^{\#}$\cite{Lee2017} & $52.72\pm1.36^{*}$  \\
	IMM$^{\#}$\cite{Lee2017} & $94.12\pm0.27^{*}$ \\
	\textbf{OWM}$^{\#}$ & $96.59\pm0.06$  \\
	SGD$^{\dagger}$ & $53.85\pm0.14^{*}$ \\
	CAB$^{\dagger}$\cite{He2018overcoming} & $94.91\pm0.30^{*}$ \\
	\textbf{OWM}$^{\dagger}$ & $96.30\pm0.03$\\
	\hline
\end{tabular}
	\caption{\textbf{Comparison of performance of different methods in disjoint MNIST tasks.}
		Network size: $^{\dagger}$, 3-layer networks with [784-800-10] neurons; $^{\#}$, 4-layer networks with [784-800-800-10] neurons. Performance results from other methods were adopted from previous studies. $^{*}$, $p<0.01$. }
	\label{performance of Disjoint MNIST}
\end{table}

\begin{table}[!h]
	\centering
	\begin{tabular}{c|c}
		\multicolumn{2}{c}{\textbf{Disjoint CIFAR10 Experiment}} \\
		\hline
		\textbf{Methods} & \textbf{Accuracy (\%)}\\
		\hline
		EWC\cite{hu2018overcoming} & $31.09$  \\
		IMM\cite{hu2018overcoming} & $32.36$ \\
		MA \cite{hu2018overcoming} & $40.47$  \\
		\textbf{OWM} & $52.83$  \\
		\hline
	\end{tabular}
	\caption{\textbf{Comparison of performance of different methods in disjoint CIFAR-10 task.} See Methods for details. MA: Model Adaptation 
   }
	\label{performance of CIFAR10}
\end{table}

In the results mentioned above, feature extractors pre-trained by the complete training sets in corresponding tasks were used to provide the feature vectors for the OWM-trained classifier. We next examined whether the classifier can learn categories on which  the feature extractor has not been trained. Results were in the affirmative, as shown in Fig.~\ref{OWM in CHW}b. For example, the feature extractor trained with 500 randomly selected Chinese characters (out of 3,755, less than $15\%$ of categories) could already support the classifier to sequentially learn the remaining 3,255 characters with near $80\%$ accuracy (chance level of $1/3,255$), demonstrating that the network could sequentially learn new categories not previously encountered. \textcolor{black}{However, we note that a higher degree of pre-taining was associated with better performance (Fig. \ref{OWM in CHW}b and c), indicating the importance of training the feature extractor on as various classes as possible.} 

Another important question is how quickly the OWM-trained classifier can learn. 
As shown in Fig. \ref{OWM in CHW}c, it only needed a small sample size to learn new mappings. 
For Chinese characters, $<10$ samples per class were sufficient to gain satisfactory performance.
\textcolor{black}{Comparison with other methods in the same task further confirmed the advantage of the OWM in achieving better performance with fewer training samples (see Supplementary Fig. 2). 
We note that the better performance achieved by the OWM with fewer samples is rooted in the well-known fact that the RLS algorithm,
from which we derived the OWM, can converge more quickly than the least mean square (LMS) algorithm, which is equivalent to the standard backpropagation \cite{Haykin2008,Shah1992}.}

\textcolor{black}{With the dateset of Chinese characters, we also analyzed network capacity in the OWM-based continual learning.
We tested two conditions, including reducing the size of the network for a given task and increasing the number of tasks for a given network. 
We observed that network performance remained stable until its size was reduced or the number of tasks was increased to a certain value, 
after which point performance declined, indicating an approach to network capacity (Fig. \ref{OWM in CHW}d, e). 
Importantly, the changes in network performance were highly correlated with decreases in the rank of the orthogonal projector, 
which is consistent with our theoretical analysis regarding network capacity in OWM-based continual learning (see Methods for details).}

\textcolor{black}{In the experiments with Chinese characters, it is possible that although a class was never seen by the network, it shared features with other classes used in feature extractor pre-training. Thus, to further test the ability of the OWM in continual learning without a pre-trained feature extractor, we examined its performance in the disjoint CIFAR-10 task. In this task, the network was trained to recognize two classes each time; thus, in a total of five consecutive tasks it learned to recognize all 10 classes. Importantly, the whole network, including both the feature extractor and classifier, was trained continually in an end-to-end manner. In this task, the OWM outperformed other recently proposed continual learning methods by a large margin (Table \ref{performance of CIFAR10}), exhibiting great potential to improve the networks' ability to learn new classes ``on the go"  and with the feature extractor free of pre-training. 
Although the performance in an end-to-end training setting was still inferior than that with a pre-trained feature extractor,  this is an important step towards removing the usual distinction between the training and application phases of DNNs, thus allowing efficient online learning. 
We note that a continual learning method for classifiers with a pre-trained feature extractor may also be useful. While the number of features in a given domain (e.g., human faces) is usually limited, the possible ways of combining different features to form a new object (e.g., individual faces) are almost infinite. 
Thus, given feature extractors pre-trained on sufficiently diverse sample sets, a classifier could greatly benefit from continual learning to recognize countless new classes.}

\section{Context Dependent Processing Module}

\begin{figure}[!h]
	\centering
	\includegraphics[width=1.0\textwidth] {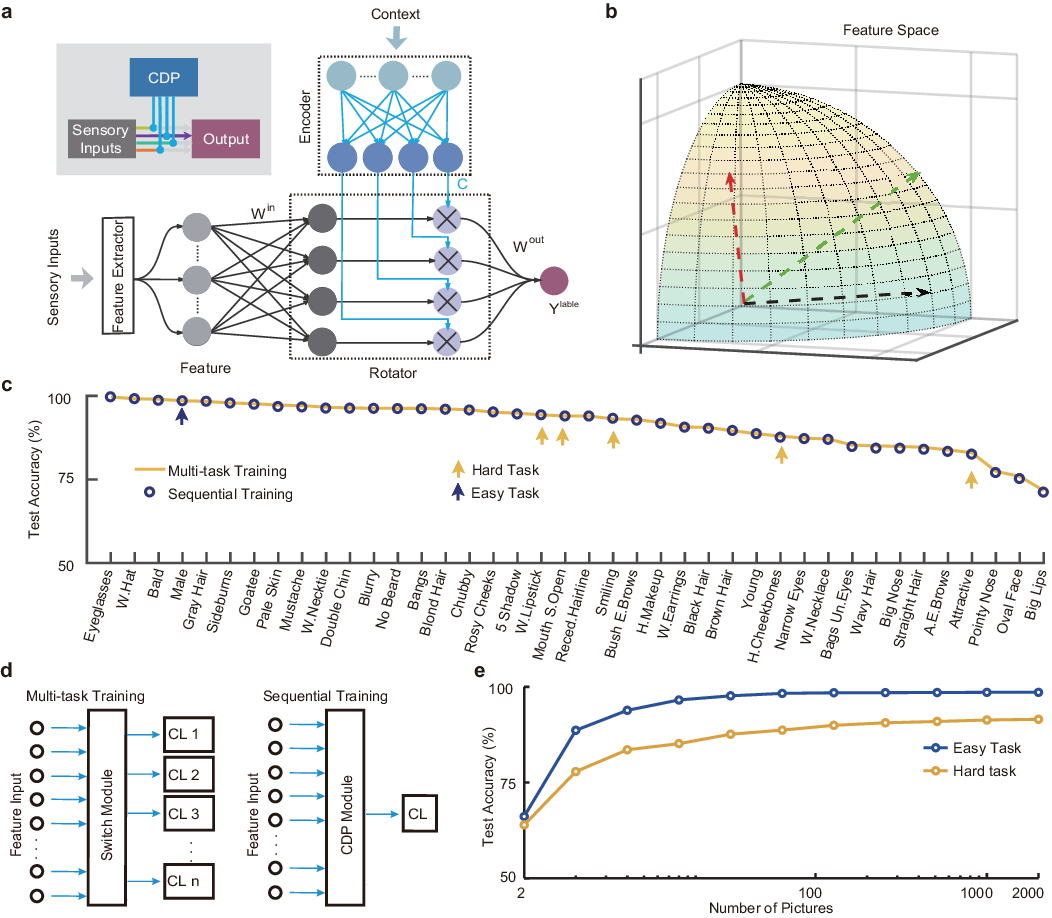} 
	\caption{ \textbf{Achieving context-dependent sequential learning via the OWM algorithm and the CDP module.}
		\textbf{a}, Schematic diagram of network architecture. The CDP module dynamically modulates the mapping between sensory inputs to network outputs according to the contextual information. The main figure and the inset illustrate the detailed internal structure of the module and the overall architecture, respectively. 
		\textbf{b}, Schematic diagram showing the role of the CDP module in rotating inputs in feature space (see Methods for details).
		\textbf{c}, Performance of sequentially learning to classify faces by 40 different attributes, each associated with a unique contextual signal, compared with results obtained by multi-task training. Tasks were sorted by test accuracy. 
		\textbf{d}, Schematic diagrams showing network architecture for multi-task (left) and sequential (right) training. CL, classifier. To achieve context-dependent processing, in multi-task training a switch module and $n$ classifiers are needed, where $n$ is the number of different attributes.
		\textbf{e}, Classification accuracies for a relatively easy task (gender; blue curve) and five more difficult, sequentially learned tasks (e.g., attractiveness; orange curve; mean results across all five tasks are shown) are plotted as a function of training sample size. Tasks and corresponding performance obtained by training on the full dataset are marked with arrows in \textbf{c}. }
	\label{OWM in celebA}
\end{figure}

Although a system that can learn many different mapping rules in an online and sequential manner is highly desirable, such a system cannot accomplish context-dependent learning by itself. To achieve that, contextual information needs to interact with sensory information properly. Here we adopted a solution inspired by the primate PFC. The PFC receives sensory inputs as well as contextual information, which enables it to choose sensory features most relevant to the present task to guide action\cite{Mante2013,Siegel2015,Fuster2015}. To mimic this architecture, we added the context-dependent processing (CDP) module before the OWM-trained classifier, which was fed with both sensory feature vectors and contextual information (Fig.~\ref{OWM in celebA}a). The CDP module consists of an encoder sub-module, which transforms contextual information to proper controlling signals, and a ``rotator" sub-module, which uses controlling signals to manipulate the processing of sensory inputs. \textcolor{black}{The encoder sub-module is trainable and learns in a continual way with the OWM.} Mathematically, the context-dependent manipulation serves by rotating the sensory input space according to the contextual information (Fig.~\ref{OWM in celebA}b, see Methods), thereby changing the representation of sensory information without interfering with its content. The rotation of the input space allows for OWM to be applied for identical sensory inputs in different contexts. To demonstrate the effectiveness of this CDP module, we trained the system to classify a set of faces according to 40 different attributes \cite{Liu2015celebA}, i.e., to learn 40 different mappings sequentially with the same sensory inputs. The contextual information was the embedding vectors \cite{Radim2012} of the corresponding task names, which were projected to control the rotation of the sensory inputs. As shown in Fig.~\ref{OWM in celebA}c, the system sequentially learned all 40 different, context-specific mapping rules with a single classifier. The accuracy was very close to that achieved by multi-task training, in which the network was trained to classify all 40 attributes using 40 separate classifiers (Fig.~\ref{OWM in celebA}d). \textcolor{black}{In addition, similar to the results obtained in learning Chinese characters, the network was able to learn context-dependent processing quickly. Here, $\sim$20 faces were enough to reach the learning plateau for both simple, e.g., male vs. female, and difficult, e.g., attractive vs unattractive, tasks (Fig.~\ref{OWM in celebA}e). 
In the experiment, our approach achieved better performance with fewer samples in comparison with other methods for continual learning, indicating its potential to enable a system to adapt quickly in highly dynamic environments with regularities changing with contexts (Supplementary Fig. 2b)}.  \textcolor{black}{Interestingly, we found that the CDP module was able to identify the meaningful signal from the contextual inputs with noise (Supplementary Fig. 3 and Supplementary Table 4, 
see Methods for task details) and to learn how to use the contextual information effectively (Supplementary Fig. 4). 
These results indicate that our approach allows the system to infer the correct context signal from experience and use it properly. Importantly, such an ability would open the door for an intelligent agent to explore environments and gradually learn its regularities in an autonomous way.}

\section*{Discussion}

If we view traditional DNNs as powerful sensory processing modules, the current approach could be understood as adding a flexible  cognitive module to the system. This architecture was inspired by the primate brain. For example, the primate visual pathway is dedicated to analyzing raw visual images  and eventually representing $\sim100$ features in higher visual areas such as the inferotemporal cortex \cite{Lehky2014}. The outputs of this “feature extractor” are then sent to the prefrontal cortex (PFC) for object identification and categorization \cite{Freedman2001Categorical,Hung2005,Kravitz2013}. The training of the feature extractor is difficult and time-consuming. In humans, it takes   years or even decades for higher visual cortices to become fully developed and reach peak performance  \cite{Gomez2017}. However, with sufficiently developed visual cortices, humans can quickly learn new visual object categories, often by seeing just a few positive examples  \cite{Xu2007}. By adding a cognitive module supporting continual learning to DNN-based  feature extractors, we found a qualitatively similar behavior in neural networks. That is, although the training of the feature extractor is computationally difficult and requires a large number of samples, with a well-trained feature extractor, the learning of new categories can be achieved quickly. This suggests that the mechanisms underlying fast concept formation in humans may be understood, at least in part, from a connectionist perspective. In addition to the role of supporting the fast learning of new concepts, another function of the primate PFC is to represent contextual information \cite{Miller2001} and use it to select those sensory features most relevant for the current task \cite{Mante2013}. This gives rise to the flexibility exhibited in primate’ behavior and here we demonstrated that similar architecture can do the same in artificial neural networks. Interestingly, we found that in the CDP module, the neuronal responses showed mixed selectivity to sensory features, contexts, and their combinations (Supplementary Fig. 5), similar to that found for real PFC neurons	\cite{Rigotti2013}. Thus, it would be informative to see whether the rotation of input space adopted in our CDP module captures the operation carried out in the real PFC. {For tasks similar to the face classification tested above, one possible solution to achieve context-dependent processing is to add additional classifier outputs for each new task/context. However, this approach only works if there is no hidden layer between the feature extractor and final output layer. Otherwise the shared weights between different classifier outputs will suffer from catastrophic forgetting during continual learning, especially if the inputs are the same for all contexts. More importantly, adding additional classifier outputs (and all related weights) for each new task/context would lead to increasingly complex and bulky systems (cf. Fig. \ref{OWM in celebA}d left). As the total number of possible contexts can be arbitrarily large, such a solution is clearly not scalable. As the total number of possible contexts can be arbitrarily large, such a solution is clearly not scalable. Finally, for artificial intelligence systems, the importance of the CDP-module would depend on application. In scenarios in which a compact system needs to learn numerous contexts "on the go", similar to what human individuals need to do within their lifetimes, the ability of the  OWM-empowered CDP-module to reuse classifiers is of paramount importance.}

As demonstrated in the present results, an efficient and scalable algorithm of continual learning is not only crucial for achieving flexible context-dependent processing, but also important to ensure, more generally, that the added cognitive module is able to learn new tasks when encountered. In continual learning, preserving previously acquired knowledge while maintaining plasticity for subsequent learning is the key \cite{Parisi2018}. In the brain, the separation of synapses utilized for different tasks is essential for sequential learning \cite{Cichon2015}, which inspired the development of algorithms to protect the important weights involved in previously learned tasks while training the network for new ones \cite{Kirkpatricka2017,Zenke2017}. However, these ``frozen" weights necessarily reduce the degrees of freedom of the system, i.e., they decrease the volume of parameter space to search for a configuration that can satisfy both old and new tasks. Here, by allowing the ``frozen" weights to be adjustable again without erasing acquired knowledge, the OWM exhibited clear advantages in performance. However, further studies are required to investigate whether algorithms similar to the OWM are implemented in the brain. {Recently, it has been suggested that a variant of backpropagation algorithm, i.e., the ``conceptor-aided back-prop" (CAB) can be used for continual learning by shielding gradients against degradation of previously learned tasks \cite{He2018overcoming}. By providing more effective shielding of gradients through constructing an orthogonal projector, the OWM achieved much better protection of previously acquired knowledge, yielding highly competitive results in empirical tests compared with the CAB (see Tables \ref{performance of Shuffled MNIST}, \ref{performance of Disjoint MNIST}, Fig. \ref{mnist_Disjoint_Task10} and Supplementary Information for details). The OWM and continual learning methods mentioned above are regularization approaches \cite{Parisi2018}. Similar to other methods within this category, the OWM exhibits tradeoff between the performance of the old and new tasks, due to limited resources to consolidate the knowledge of previous tasks. In contrast to regularization approaches, other types of continual learning methods involve dynamically introducing extra neurons or layers along the learning process \cite{Rusu2016}, which may help mitigate the tradeoff  described above \cite{Parisi2018}. However, regularization approaches require no extra resources to accommodate newly acquired knowledge during training and, therefore,  are capable of producing compact yet versatile systems.}

We note that a solution for continual learning based on context-dependent processing was suggested  recently \cite{masse2018alleviating}. 
In this work, a context-dependent gating mechanism was used to separate subnetworks for processing individual tasks during continual learning. 
However, for this approach to work, the same contextual information needs to be present during both the training and testing phases. 
As such information is rarely available in practical situations, 
this seriously limits the applicability of the method. 
Different from this approach, the CDP module in our work enables the network to modulate its processing according to the contextual information so that the same inputs can be treated differently in different contexts. 
This role is not related to continual learning, as the CDP module is needed in the same task as shown in Fig. \ref{OWM in celebA}, even if the system was trained concurrently. Importantly, as contextual information, e.g., environmental cues, the task at hand, etc. is always available for any input that needs context-dependent processing, 
the limitations of using context-dependent gating for continual learning is not a problem for our CDP module.

{Other biologically inspired approaches for continual learning are based on complementary learning systems (CLS) theory \cite{McClelland1995, Kumaran2016}. Such systems involve interplay between two sub-systems similar to the mammalian hippocampus and neocortex, i.e., a task-solving network (neocortex) accompanied by a generative network (hippocampus) to maintain the memories of previous tasks \cite{Shin2017}. With the aid of the Learning without Forgetting (LwF) method \cite{Li2017LwF}, data for old tasks sampled by the generative module are interleaved with those for the current task to train the neural network  to avoid catastrophic forgetting. Although here we used a completely different approach for continual learning, the CLS framework may also be instrumental for further development of our approach.
\textcolor{black}{Currently, the encoder of the CDP module has the ability to infer contextual information from the environment and also to learn how to use it effectively. Conceivably it could be further developed to recognize and classify complex contexts. Such a flexible module for recognizing proper contextual signals may be analogous to the hippocampus in the brain, which is related to the classification of different environmental cues via pattern separation and completion \cite{Kumaran2016}. Thus, it would be informative for future studies to investigate whether the current approach can be combined with the CLS framework to achieve more flexible and sophisticated context-dependent processing. }	
	
Taken together, our study demonstrated that it is possible to teach a highly compact network many context-dependent mappings sequentially. Although we demonstrated its effectiveness here with the supervised learning paradigm, the OWM has the potential to be applied to other training frameworks. {Another regularization approach for overcoming catastrophic forgetting, i.e., the EWC, has been successfully implemented in reinforcement learning \cite{Kirkpatricka2017}. As the EWC can be viewed as a special case of OWM in some circumstances (see Supplementary Information for details), it suggests that similar procedures could be extended for the use of the OWM and CDP module in unsupervised conditions, thereby enabling networks to learn different mapping rules for different contexts by reinforcement learning. We expect that such an approach, combined with effective methods of knowledge transfer, e.g., \cite{Rohrbach2010,yosinski2014transferable,Hinton2015, Schwarz2018}, may eventually lead to systems with sufficient flexibility to work in complex and dynamic situations}.
	
\newpage
\section*{Methods}
\noindent\textbf{The OWM algorithm}. Consider a feed-forward network of $L+1$ layers, indexed by $l=0,1,\cdots,L$ with $l=0$ and  $l=L$ being the input and output layer, respectively. All hidden layers share the same activation function $g(\bullet )$.  ${{\mathbf{W}}_{l}}$ represents the connections between the $(l-1)$th and $l$th layer with ${{\mathbf{W}}_{l}}\in {{\mathbb{R}}^{s\times m}}$. $\mathbf{x}_l$ and $\mathbf{y}_l$ denote the output and input of the $l$th layer, respectively, where $\mathbf{x}_l=g(\mathbf{y}_l)$ and $\mathbf{y}_l=\mathbf{W}_l^T\mathbf{x}_{l-1}$. ${{\mathbf{x}}_{l-1}}\in {{\mathbb{R}}^{s}}$ and  ${{\mathbf{y}}_{l}}\in {{\mathbb{R}}^{m}}$.
	
In the OWM, the orthogonal projector $\mathbf{P}_l$ defined in the input space of layer $l$ for learned tasks is key for overcoming catastrophic interference in sequential learning. In practice, $\mathbf{P}_l$ can be recursively updated for each task in a way similar to calculating the correlation-inverse matrix $\mathbf{P}^{(RLS)}=({\sum_{i=1}^n}\mathbf{x}(i)\mathbf{x}^T(i)+\alpha \mathbf{I})^{-1}$ in the RLS algorithm \cite{Haykin2008,Singhal1989,Shah1992} (see discussion on relationship of OWM and RLS in Supplementary Information). This method allows $\mathbf{P}_l$ to be determined based on the current inputs and the $\mathbf{P}_l$ for the last task. It also avoids matrix-inverse operation in the original definition of $\mathbf{P}_l$.
	
Below we provide the detailed procedure for the implementation of the OWM method.
\begin{itemize}
		\item[] a. Initialization of parameters: randomly initialize $\mathbf{W}_l(0)$ and set ${{\mathbf{P}}_{l}}(0)= {{\mathbf{I}}_{l}}/\beta$ for $l=1,\cdots,L$.
		\item[]    b. Forward propagate the inputs of the $i$th batch in the $j^{th}$ task, then back propagate the errors and calculate weight modifications $\Delta \mathbf{W}_{l}^{BP}(i,j)$ for ${{\mathbf{W}}_{l}}(i-1,j)$ by the standard BP method.
		\item[] c. Update the weight matrix in each layer by
		\begin{equation}\label{updateW}
		\begin{aligned}
		{{\mathbf{W}}_{l}}(i,j)&={{\mathbf{W}}_{l}}(i-1,j)+\kappa (i,j)\Delta \mathbf{W}_{l}^{BP}(i,j)\quad\mbox{if}\quad j=1\\
		{{\mathbf{W}}_{l}}(i,j)&={{\mathbf{W}}_{l}}(i-1,j)+\kappa (i,j){{\mathbf{P}}_{l}}(j-1)\Delta \mathbf{W}_{l}^{BP}(i,j)\quad\mbox{if}\quad j=2,3,\cdots
		\end{aligned}
		\end{equation}
		where $\kappa (i,j)$ is the predefined learning rate.
		\item[] d. Repeat steps (b) to (c) for the next batch.
		\item[] e. If the $j^{th}$ task is accomplished, forward propagate the mean of the inputs for each batch
		($i=1,\cdots,{{n}_{j}}$) in the $j^{th}$ task successively. Update ${{\mathbf{P}}_{l}}$ for ${{\mathbf{W}}_{l}}$ as $\mathbf{P}_l(j)=\mathbf{P}_l(n_j,j)$, where $\mathbf{P}_l(j)=\mathbf{P}_l(n_j,j)$ can be calculated iteratively according to:
		\begin{equation}\label{EqRls}
		\begin{aligned}
		{{\mathbf{P}}_{l}}(i,j) &={{\mathbf{P}}_{l}}(i-1,j)-{{\mathbf{k}}_{l}}(i,j){{{\mathbf{\bar{x}}}}_{l-1}}{{(i,j)}^{T}}{{\mathbf{P}}_{l}}(i-1,j) \\
		{{\mathbf{k}}_{l}}(i,j) &={{\mathbf{P}}_{l}}(i-1,j){{{\mathbf{\bar{x}}}}_{l-1}}(i,j)
		/[\alpha+{{{\mathbf{\bar{x}}}}_{l-1}}{{(i,j)}^{T}}{{\mathbf{P}}_{l}}(i-1,j){{{\mathbf{\bar{x}}}}_{l-1}}(i,j)] \\
		\end{aligned}
		\end{equation}
		in which ${{\mathbf{\bar{x}}}_{l-1}}(i)$ is the output of the $l-1^{th}$ layer in response to the mean of the inputs in the $i^{th}$ batch of the$j^{th}$ task, and $\mathbf{P}_l(0,j)=\mathbf{P}_l(j-1)$.
		\item[] f. Repeat steps (b) to (e) for the next task.
		
\end{itemize}
	
We note that the algorithm achieved the same performance if the orthogonal projector $\mathbf{P}_l$ was updated for each batch according to Eq. \ref{EqRls}, with $\alpha$ decaying as $\alpha_{i,j}=\alpha_0\lambda^{i/n_j}$ for the $i$th batch of data in the $j$th task. This method can be understood as treating each batch as a different task. It avoids the extra storage space as well as data reloading in (d) and, therefore, significantly accelerates processing. In this case, if the learning rate is set to $\kappa (i)=1/[1+{{\mathbf{\bar{x}}}_{l-1}}{{(i)}^{T}}{{\mathbf{P}}_{l}}(i-1){{\mathbf{\bar{x}}}_{l-1}}(i)]$ and $\alpha_{i,j}$ is permanently set to $\alpha_{i,j}=1$, the procedure essentially uses RLS to train the neural network under the name of Enhanced Back Propagation  (EBP), which is proposed to increase the speed of convergence in training \cite{Shah1992}. Therefore, our algorithm has the same computational complexity as EBP---O($N_nN_w^2$), where $N_n$ is the total number of neurons and $N_w$ is the number of input weights per neuron \cite{Shah1992}.
	
In addition, we analyzed the capacity of the OWM, i.e., how many different tasks could be learned using this method. The capacity of one network layer can be measured by the rank of $\mathbf{P}_i$, which is defined as the orthogonal projector calculated after task $i$, with  $\Delta \mathbf{P}_{i+1}$ then defined as the update in the next task satisfying $\mathbf{P}_{i+1} =\mathbf{P}_i - \Delta \mathbf{P}_{i+1} $.  As $range(\mathbf{P}_{i+1} )\cap range(\Delta \mathbf{P}_{i+1} )=\varnothing $, $rank(\mathbf{P}_{i+1} )=rank(\mathbf{P}_i )- rank(\Delta \mathbf{P}_{i+1} )$. In the ideal case where each task consumes the capacity effectively, as the learning process continues, the rank of $\mathbf{P}_l$ is approaching 0, indicating that this particular layer no longer has the capacity to learn new tasks.  \textcolor{black}{The capacity of the whole network can be  approximated by the summation of the capacity of each layer: $rank_{tot}=\sum_{l=1}^L rank(\mathbf{P}_l)/rank(\beta\mathbf{I})$ where $\beta\mathbf{I}$ is the initial value of matrix $\mathbf{P}$. The rank is normalized to balance the contribution of each layer}. 
We conducted two experiments (Fig. \ref{OWM in CHW}d,e)  on the CASIA-HWDB1.1 dataset to verify the above analysis. 
In the experiments, to avoid influence by the  tolerance value in the calculation of matrix rank, the rank was estimated as $rank(\mathbf{P})=\sum_{i=1}s_i(\mathbf{P})/\beta$, 
where $s_i(\bullet)$ denotes the $i$th singular value of the matrix.   }
If the capacity limit of the entire network is finally approached, two solutions can be considered: 1) introduction of a larger $\alpha$ or the forgetting factor used in RLS \cite{Haykin2008} and online EWC ~\cite{Schwarz2018}; and 2) addition of more layer(s), e.g., CDP module (see below for details), to provide more space to preserve previously learned knowledge.
	
\noindent\textbf{The CDP module}. In context-dependent learning,  to change the representation of sensory inputs without distorting information content in different contexts, we added one layer of neurons after the output layer of the feature extractor (cf. Fig. \ref{OWM in celebA}a). Below we describe, from a mathematical point of view, how this CDP layer works, using the face classification task as an example.
	
In this task, the rotator sub-module was fed with feature vectors for different faces,
	$\mathbf{F}={{[{{f}_{1}},{{f}_{2}}\cdots ,{{f}_{k}}]}^{\text{T}}}\in {{\mathbb{R}}^{k}}$,
	and modulated by  non-negative controlling signals, $\mathbf{C}={{[{{c}_{1}},{{c}_{2}},...,{{c}_{m}}]}^{T}}\in {{\mathbb{R}}^{m}}$.
	The controlling signals $\mathbf{C}$ were drawn from the contextual information (word vector of corresponding task name) by the encoder sub-module.
	Then the CDP module outputted  ${{\mathbf{Y}}^{\text{out}}}={{[{{y}_{1}},{{y}_{2}},...,{{y}_{m}}]}^{\text{T}}}\in {{\mathbb{R}}^{m}}$,
	with ${{y}_{i}}={{c}_{i}}\operatorname{g}( (\mathbf{w}_{i}^{in})^{T}\mathbf{F} ) $, to a classifier for further processing.
	The input weight ${{\mathbf{W}}^{\text{in}}}=[\mathbf{w}_{1}^{\text{in}},\mathbf{w}_{2}^{\text{in}},..,\mathbf{w}_{m}^{\text{in}}]\in {{\mathbb{R}}^{k\times m}}$
	of the CDP module was randomly initialized and fixed across all contexts. 
	The rest weights in the CDP module, including the output weight ${{\mathbf{W}}^{\text{out}}}$ and the weights in the encoder, were trained by the OWM method.  
The function of the CDP module can then be summarized as
\begin{equation}\label{EqPFC}
\begin{aligned}
{{\mathbf{Y}}^{\text{out}}}&=\operatorname{g}\left( {{\left( {{\mathbf{W}}^{\text{in}}} \right)}^{\text{T}}}\mathbf{F} \right)\odot \mathbf{C} \\
&=\operatorname{g}\left( {{\left[ \begin{matrix}
	\mathbf{w}_{1}^{\text{in}} & \mathbf{w}_{2}^{\text{in}} & \cdots  & \mathbf{w}_{m}^{\text{in}}  \\
	\end{matrix} \right]}^{\text{T}}}\mathbf{F} \right)\odot \mathbf{C}  \\
&=\operatorname{g}\left( {{\left[ \begin{matrix}
	{{c}_{1}}\left\| \mathbf{F} \right\|\left\| \mathbf{w}_{1}^{\text{in}} \right\|cos{{\theta }_{1}}, & {{c}_{2}}\left\| \mathbf{F} \right\|\left\| \mathbf{w}_{2}^{\text{in}} \right\|cos{{\theta }_{2}}, & \cdots , & {{c}_{m}}\left\| \mathbf{F} \right\|\left\| \mathbf{w}_{m}^{\text{in}} \right\|cos{{\theta }_{m}}  \\
	\end{matrix} \right]}^{\text{T}}} \right) \\
&=\operatorname{g}\left( {{\left[ \begin{matrix}
	{{c}_{1}}\left\| \mathbf{w}_{1}^{\text{in}} \right\|cos{{\theta }_{1}}, & {{c}_{2}}\left\| \mathbf{w}_{2}^{\text{in}} \right\|cos{{\theta }_{2}}, & \cdots , & {{c}_{m}}\left\| \mathbf{w}_{m}^{\text{in}} \right\|cos{{\theta }_{m}}  \\
	\end{matrix} \right]}^{\text{T}}} \right)\left\| \mathbf{F} \right\|\  \\
\end{aligned}
\end{equation}
where $\odot$ represents element-wise multiplication and ${{\theta }_{i}}$ is the angle between $\mathbf{w}_{i}^{\text{in}}$ and $\mathbf{F}$.  Note that for any $\upsilon \ge 0$,  $\operatorname{g}(\upsilon x)=\text{max}(0,\upsilon x)=\upsilon \text{max}(0,x)=\upsilon \operatorname{g}(x)$. 
\textcolor{black}{ The ReLU function was used in the current study for $g(\bullet)$ but this is not necessary. 
$g(\bullet)$ can also be chosen as a hyperbolic tangent function or logistic function.
As $\mathbf{W}^{in}$ was initialized by the Xavier method\cite{glorot2010understanding} in most cases, $\mathbf{w}_i^{in}\mathbf{F}$ was located in the linear range.
Thus the Eq.(\ref{EqPFC}) can approximately hold even for activation functions other than ReLU. 
We confirmed that the average accuracies for the same tasks in Fig. \ref{OWM in celebA}c with   hyperbolic tangent function (90.93\%) and logistic function (90.05\%) were close to that with ReLU ( 90.38\%). }
	
For individual faces, given the same feature vector $\mathbf{F}$ and fixed $\mathbf{W}^{in}$, $\cos {{\theta }_{i}}$ is constant. Thus, output ${{\mathbf{Y}}^{\text{out}}}$ is affected by the controlling signal $\mathbf{C}$, which is different across tasks. If we normalize $\mathbf{C}$ by $\sqrt{\sum\nolimits_{i=1}^{m}{{{\left( {{c}_{i}}\left\|\mathbf{w}_{i}^{\text{in}}\right\|\operatorname{g}\left( cos{{\theta }_{i}} \right) \right)}^{2}}}}$, it is apparent from Eq. \ref{EqPFC} that the CDP layer ``rotates" the input vector in feature space, as illustrated in Fig.~\ref{OWM in celebA}b. This explains why this added layer can change the representation of sensory inputs while keeping information contents unchanged. Importantly, it also enables the system to sequentially learn different tasks with the OWM for identical inputs.

To examine whether the CDP module can infer the correct context from distracting noise
in the environment, 
four face recognition tasks were conducted continually with the OWM, except that the explicit context signal was not presented. 
The context signals and distracting noises were simultaneously fed to the CDP module. 
The noise was sampled from a Gaussian distribution with the same mean and variance as the context signal, and varied on a trial-by-trial basis. 
In the training phase for different tasks, the position of the context signal and noises could be swapped (Supplementary Fig. 3).  
During the testing phase, either the corresponding context+noises or only noises were presented.

\noindent\textbf{Shuffled MNIST experiment.}  The shuffled MNIST experiment \cite{Goodfellow2013,Lee2017,Kirkpatricka2017,Zenke2017,He2018overcoming} usually consists of a number of sequential tasks. All tasks involve classifying handwritten digits from 0 to 9. However, for each new task, the pixels in the image are randomly shuffled, with the same randomization across all digits in the same task and different randomization across tasks. For this experiment, we trained 3- or 4- layer, feed-forward networks with [784-800-10] (3-layer) or [784-800/2000-800/2000-10] (4-layer) neurons (see Table  ~\ref{performance of Shuffled MNIST} for details) to minimize cross entropy loss by the OWM method. The ReLU activation function \cite{Nair2010} was used in the hidden layer. During training, the L2 regularization coefficient was 0.001. Dropout was applied with a drop rate of 0.2.

Table  \ref{performance of Shuffled MNIST} shows the performance of the OWM method for the shuffled MNIST tasks in comparison with other continual learning algorithms. The accuracy of the OWM method was measured by repeating the experiments 10 times. The results of other algorithms were adopted from corresponding publications. The size of the network, regarding the number of layers and number of neurons in each layer, was the same as in previous publications for a fair comparison.

Two-sided t-tests were used to compare performance between the OWM and other continual learning methods for both the shuffled and disjoint (see below) MNIST experiments. The $t$ values were calculated according to the means and standard deviations across 10 experiments. Significance was considered at $p<0.01$ with results shown in Table  \ref{performance of Shuffled MNIST}.

\noindent\textbf{Disjoint MNIST experiment.} In the 2-disjoint MNIST experiment \cite{Srivastava2013}, the original MNIST dataset was divided into two parts: The first contained digits from 0 to 4 and the second consisted of digits from 5 to 9. Correspondingly, the first task was to recognize digits among 0, 1, 2, 3 and 4 and the second task was to recognize digits among 5, 6, 7, 8 and 9. In the 10-disjoint MNIST task, 10 digits, from 0 to 9, were learned sequentially. Again, to facilitate comparison, network size and architecture were the same as in previous work \cite{Srivastava2013}.  During training, momentum optimization was applied, and the learning rate for all layers remained the same during training. Performance was calculated based on 10 repeated experiments and is shown in Table \ref{performance of Disjoint MNIST}.

\noindent\textbf{Sequential learning of classification tasks with Chinese characters and ImageNet.} Classification tasks with  ImageNet and Chinese handwritten characters are more challenging due to the complex structure in each image and more classes to ``memorize" in a sequential learning task. In sequential learning, the training for a new task started only when the neural network accomplished the current task well enough, defined here as $<1\%$ accuracy gap between two successive training epochs.   For these two tasks, we first trained a DNN as the feature extractor on the whole or partial dataset to extract features of each image.  The extracted feature vectors were then fed into a 3-layer classifier with [1024-4000-3755] neurons for the Chinese characters task and [2048-4000-1000] neurons for the ImageNet task. The classifier was trained to recognize each of the classes sequentially using the OWM method, with results shown in Supplementary Table 1. {We note that in these experiments,  as in other tests mentioned above, no negative samples were used for training the network to recognize a new class. In other words, only positive samples of a particular class were presented to the network during training.} 

\noindent\textbf{Disjoint CIFAR-10 experiment.} 
In contrast to the pre-training of feature extractors in the tasks of Chinese characters and ImageNet,
the feature extractor was trained together with the classifier in an end-to-end way using the OWM in this task. 
The CIFAR-10 dataset was divided into 5 groups. 
Each group included 2 classes of samples used to train the whole network  in one task. 
The feature extractor  consisted of three convolutional layers and the classifier consisted of three fully connected layers.    
The three convolutional layers had 64, 128, and 256 filters, respectively, 
and the size of the convolution kernel was $2\times2$. 
A maxpooling layer with size of $2\times2$ was attached to each convolutional layer. 
Dropout was applied to each maxpooling layer with a dropping probability of 0.2. The features extracted were flattened and then fed to the classifier of [1000-1000-10] neurons. 
The activation function for all layers was the ReLU function. 
The initial weights for all layers were in accordance with the Xavier initialization method proposed by Glorot and Bengio\cite{glorot2010understanding}. 
Cross entropy loss was applied for the training. 
Table \ref{performance of CIFAR10} compares the performance of the OWM with other methods using the same network structure and task. 

\noindent\textbf{Context-dependent face recognition with CelebA}. In this experiment, we first trained a feature extractor using the architecture of ResNet50 \cite{He2016Res} on the whole training dataset and with the conventional multi-task training procedure. The outputs of the feature extractor were then fed into the CDP module, which also received contextual information (cf. Fig.~\ref{OWM in celebA}a in the main text).  The rotator layer contained 5000 neurons. The size of the encoder layer was [200-5000], with ReLU applied as the activation function.  For the face classification task in the present study, rotated feature vectors were  fed directly into the classifier by weights ${{\mathbf{W}}^{\text{out}}}$. 
	Before training, all weights and biases were randomly initialized. ${{\mathbf{W}}^{\text{out}}}$ and the weights in the encoder were modified by the OWM method. Detailed results of classifying individual attributes are listed in Supplementary Table 5.

\noindent\textbf{Network parameters.} Weights in the hidden layers of the classifiers for the tasks other than disjoint CIFAR-10 task were initialized according to previously suggested method \cite{He2015actFun}. The output layers were all initialized to zero. The biases of each layer were randomly initialized according to a uniform distribution within (0, 0.1). The ReLU neurons were applied to every hidden layer in all experiments. The momentum in all optimization algorithms was  0.9.
The details of hyperparameters used for feature extractors are shown in Supplementary Table 6. Early stopping was used for training both the feature extractors and classifiers.  The hyperparameters for the OWM method are shown in Supplementary Table 7.
For tasks with MNIST and CelebA, the classifier was trained to minimize cross entropy loss, whereas for tasks with ImageNet and Chinese characters, the classifier was trained to minimize mean squared loss. \textcolor{black}{Note that cross-entropy loss was also suitable for the latter datasets. However, mean squared loss is easier to compute and less time-consuming when many tasks are involved. }

\noindent\textbf{Mixed selectivity analysis.}  For classifying different facial attributes, responses of neurons in the CDP were analyzed to examine if they exhibited mixed selectivity similar to that of real PFC neurons.
To this end, we chose two attributes with low correlation, i.e., Attractiveness (Task 1) and Smile (Task 2).  Both has about $50\%$ positive and $50\%$ negative samples in the whole dataset. The responses of each neuron in the CDP module to different inputs as well as contextual signals were analyzed with the weights in the encoder sub-module fixed for both contexts. There were 19962 test pictures, with $90\%$ correctly classified after training for both tasks. The threshold of excitation for each neuron was chosen as the average activity level across all neurons during the processing of all correctly-classified pictures. Supplementary Fig. 5 shows the selectivity of three exemplar neurons. According to the criteria usually used in electrophysiological experiments, these three neurons belonged to different categories, including task-sensitive (Neuron 1) and attribute-sensitive (Neuron 2). Importantly, Neuron 3 exhibited complex selectivity towards combinations of task and sensory attributes, as well as combinations of different attributes. This mixed selectivity is commonly reported for real PFC neurons \cite{Ramirez2016}.

\section*{Data and code availability}

All data used in this paper are publicly available and can be accessed at: \url{http://yann.lecun.com/exdb/mnist/} for the MNIST dataset; \url{https://www.cs.toronto.edu/~kriz/cifar.html} for the CIFAR dataset; \url{http://image-net.org/index} for the ILSVR2012 dataset; \url{http://www.nlpr.ia.ac.cn/databases/handwriting/Home.html} for the CASIA-HWDB dataset; \url{http://mmlab.ie.cuhk.edu.hk/projects/CelebA.html} for the CelebA dataset. For more details of the dataset, please refer to the references cited in the Dataset section of the Supplementary Information.

The source code can be accessed at \url{https://github.com/beijixiong3510/OWM}.

\bibliographystyle{naturemag}
\bibliography{owm}

\section*{Acknowledgements}

The authors thank Dr. Danko Nikoli\'c for helpful discussions. This work was supported by the National Key Research and Development Program of China (2017YFA0105203), Natural Science Foundation of China (81471368), the Strategic Priority Research Program of the Chinese Academy of Sciences (CAS) (XDB32040200), and the Hundred-Talent Program of CAS (for S.Y.).

\section*{Contributions}

S.Y., Y.C. and G.Z conceived the study and designed the experiments. G.Z. and Y.C. conducted computational experiments and theoretical analyses. C.B. assisted with some experiments and analyses. S.Y., Y.C. and G.Z. wrote the paper. 

\section*{Competing interests}

The Institute of Automation, Chinese Academy of Sciences has submitted the patent applications on the OWM algorithm (application No. PCT/CN2019/083355; invented by Chen Yang, Guanxiong Zeng and Shan Yu; pending) and the CDP module (application No. PCT/CN2019/083356; invented by Guanxiong Zeng, Chen Yang and Shan Yu; pending).

\newpage

\end{document}